\documentclass[a4paper]{article}

\usepackage{INTERSPEECH_v2}
\usepackage{comment}

\title{An End-to-End Trainable Neural Network Model with \\ Belief Tracking for Task-Oriented Dialog}
\name{} \address{} \email{}
\name{Bing Liu$^1$, Ian Lane$^1$$^,$$^2$}
\address{
 $^1$Electrical and Computer Engineering, Carnegie Mellon University\\
 $^2$Language Technologies Institute, Carnegie Mellon University}
\email{liubing@cmu.edu, lane@cmu.edu}

\begin{document}

\maketitle
\begin{abstract}

We present a novel end-to-end trainable neural network model for task-oriented dialog systems. The model is able to track dialog state, issue API calls to knowledge base (KB), and incorporate structured KB query results into system responses to successfully complete task-oriented dialogs. The proposed model produces well-structured system responses by jointly learning belief tracking and KB result processing conditioning on the dialog history. We evaluate the model in a restaurant search domain using a dataset that is converted from the second Dialog State Tracking Challenge (DSTC2) corpus. Experiment results show that the proposed model can robustly track dialog state given the dialog history. % Our model also demonstrates promising performance in per-response accuracy comparing to prior end-to-end trainable neural network models, and obtains performance comparable to state-of-the-art systems. 
Moreover, our model demonstrates promising results in producing appropriate system responses, outperforming prior end-to-end trainable neural network models using per-response accuracy evaluation metrics.

\end{abstract}
\noindent\textbf{Index Terms}: spoken dialog systems, end-to-end model, task-oriented, dialog state tracking, language understanding

\section{Introduction}
Task-oriented spoken dialog system is a prominent component in today's virtual personal assistants, which enable people to perform everyday tasks by interacting with devices via voice input. Traditional task-oriented dialog systems have complex pipelines, with a number of independently developed and modularly connected components. There are usually separated modules in a pipeline for natural language understanding (NLU), dialog state tracking (DST), dialog management (DM), and natural language generation (NLG) \cite{rudnicky1999creating,young2006using,raux2005let,young2013pomdp}. One limitation with such pipeline approach is that it is inherently hard to adapt a system to new domains, as all these modules are trained and fine-tuned independently. Moreover, errors made in upper stream modules may propagate to downstream components, making it tedious to identify and track the source of error \cite{zhao2016towards}.

% To address these limitations, efforts have been made recently in designing end-to-end frameworks for task-oriented dialogs. Wen et al. \cite{wenN2N16} proposed an end-to-end trainable neural network model with modularly connected neural networks for each system component. Zhao and Eskenazi \cite{zhao2016towards} introduced an end-to-end reinforcement learning framework that jointly performs dialog state tracking and policy learning. Li et al. \cite{li2017end} proposed an end-to-end learning framework that leverages both supervised and reinforcement learning signals and showed promising dialog modeling performance. Bordes and Weston \cite{bordes2016learning} proposed an end-to-end memory network method, and modelled task-oriented dialog with a reasoning approach without explicitly learning the dialog policy and tracking belief state. 

To address these limitations, efforts have been made recently in designing end-to-end frameworks for task-oriented dialogs. Wen et al. \cite{wenN2N16} proposed an end-to-end trainable neural network model with modularly connected neural networks for each system component. Zhao and Eskenazi \cite{zhao2016towards} introduced an end-to-end reinforcement learning framework that jointly performs dialog state tracking and policy learning. Li et al. \cite{li2017end} proposed an end-to-end learning framework that leverages both supervised and reinforcement learning signals and showed promising dialog modeling performance. Such end-to-end trainable neural network models can be optimized directly towards the final system objective functions (e.g. task success rate) and thus ameliorate the challenges of credit assignment and online adaptation \cite{zhao2016towards}. 

In this work, we present an end-to-end trainable neural network model for task-oriented dialog that applies a unified network for belief tracking, knowledge base (KB) operation, and response creation. The model is able to track dialog state, interface with a KB, and incorporate structured KB query results into system responses to successfully complete task-oriented dialogs. We show that our proposed model can robustly track belief state given the dialog history. Our model also demonstrates promising performance in providing appropriate system responses and conducting task-oriented dialogs compared to prior end-to-end trainable neural network models.

        \begin{figure*}[t]
          \centering
          \includegraphics[width=390pt]{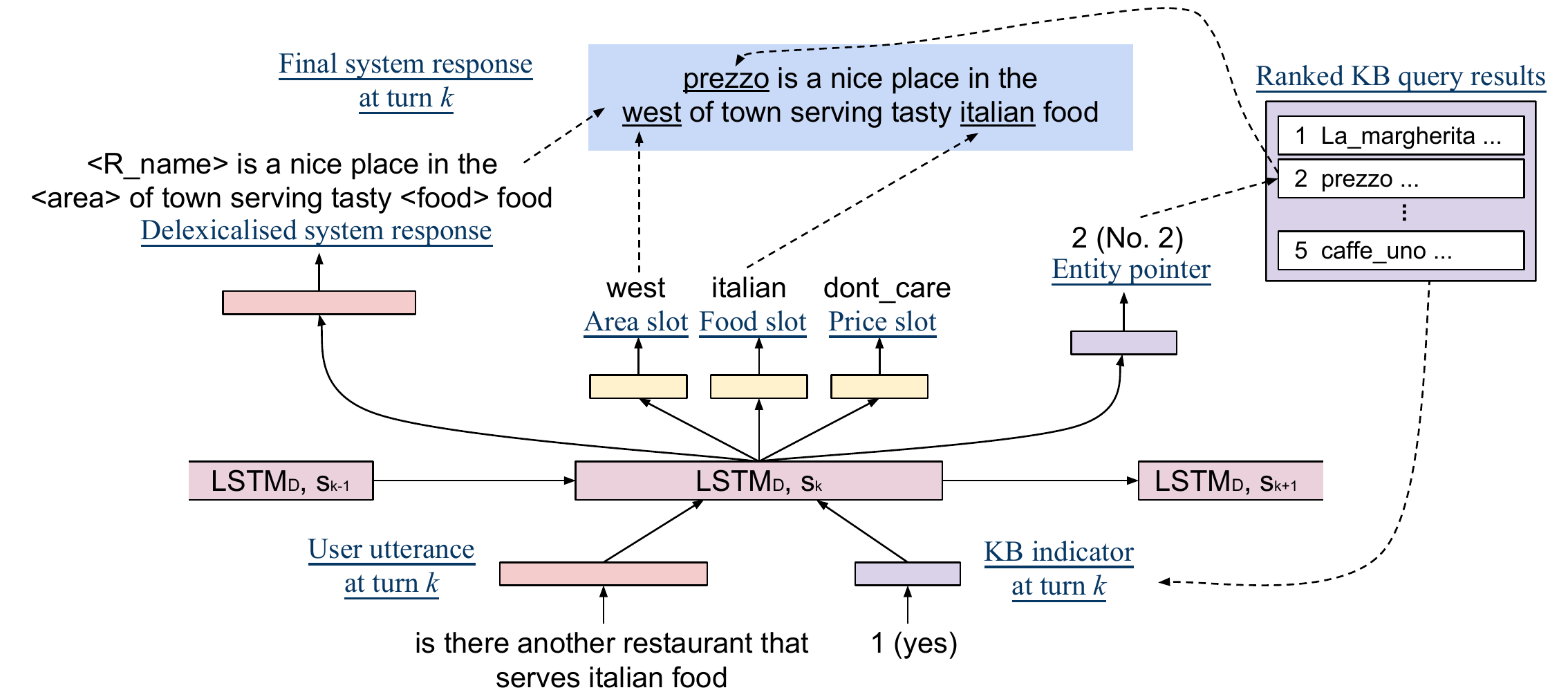}
          \caption{System architecture of the proposed end-to-end trainable neural network model for task-oriented dialog.}
          \label{fig:system_diagram_v3}
        \end{figure*}  
        
\section{Related Work}

\subsection{Dialog State Tracking}
In spoken dialog systems, dialog state tracking, or belief tracking, refers to the task of maintaining a distribution over possible dialog states which directly determine the system’s actions. Dialog state tracker is a core component in many state-of-the-art task-oriented spoken dialog systems \cite{wenN2N16,williams2016dialog}. Conventional approaches for DST include using rule-based systems and generative methods that model the dialog as a dynamic Bayesian network \cite{williams2007partially}. Discriminative approaches using sequence models such as CRF \cite{lee2013structured} or RNN \cite{henderson2014robust,henderson2014word} address the limitation of generative models with the flexibility in exploring arbitrary features \cite{henderson2015machine} and achieve state-of-the-art DST performance. 

\subsection{End-to-End Task-Oriented Dialog Models}
Conventional task-oriented dialog systems typically require a large number of domain-specific rules and handcrafted features, which make it hard to extend a good performing model to new application domains. Recent approaches to task-oriented dialog cast the task as a partially observable Markov Decision Process (POMDP) \cite{young2013pomdp} and use reinforcement learning for online policy optimization by interacting with users \cite{gavsic2013line}. The dialog state and system action spaces have to be carefully designed in order to make the reinforcement policy learning tractable \cite{young2013pomdp}. 

With the success of end-to-end trainable neural network models in non-task-oriented chit-chat dialog settings \cite{shang2015neural,serban2015building}, efforts have been made in carrying over the good performance of end-to-end trainable models to task-oriented dialogs. Wen et al. \cite{wenN2N16} proposed a neural network based model that is end-to-end trainable yet still modularly connected. The model has separated modules for intent estimation, belief tracking, policy learning, and response generation. Our model, on the other hand, use a unified network for belief tracking, KB operation, and response generation, to fully explore knowledge that can be shared among different tasks. Bordes and Weston \cite{bordes2016learning} recently proposed modeling the dialog with a reasoning approach using end-to-end memory network. Their model selects a best system response directly from a list of response candidates without explicitly tracking dialog state. Comparing to this approach, our model tracks dialog state over the sequence of turns explicitly, as it is shown in \cite{jurvcivcek2012reinforcement} that robust dialog state tracking is likely to boost success rate in task completion. Moreover, when generating final system response, instead of letting the model to select a final response directly from a large pool of candidate responses, we let our model to select skeletal sentence structure from a short list of candidates and then replace the delexicalised tokens with the state tracking outputs. This method will help to reduce the number of training samples required \cite{henderson2014robust} and make the model more robust to noises in dialog state.

\section{Proposed Method}
    We model task-oriented dialog as a multi-task sequence learning problem, with components for encoding user input, tracking belief state, issuing API calls, processing KB results, and generating system responses. The model architecture is as shown in Figure 1. Sequence of turns in a dialog is encoded using LSTM \cite{hochreiter1997long} recurrent neural networks. Conditioning on the dialog history, state of the conversation is maintained in the LSTM state. The LSTM state vector is used to generate: (1) a skeletal sentence structure by selecting from a list of delexicalised system response candidates, (2) a probability distribution of values for each slot in belief tracker, and (3) a pointer to an entity in the retrieved KB results that matches the user's query. The final system response is generated by replacing the delexicalised tokens with the predicted slot values and entity attribute values. Each model component is described in detail in below sections.
   
\subsection{Utterance Encoding}
    Utterance encoding here refers to encode a sequence of words into a continuous dense vector. Popular methods include using bag-of-means on word embeddings and RNNs \cite{liu2016attention,hakkani2016multi}. We use bidirectional LSTM to encode the user input to an utterance vector. Let $\mathbf{U}_k = (w_1, w_2, ..., w_{T_k})$ be the user input at the $k$th turn with $T_k$ words. The user utterance vector $U_k$ is represented by: $U_k =  [\overrightarrow{h^{U_k}_{T_k}}, \overleftarrow{h^{U_k}_{1}}]$,
%        \begin{align}
%            & U_k =  [\overrightarrow{h^{U_k}_{T_k}}, \overleftarrow{h^{U_k}_{1}}]       
%        \end{align}
    where $\overrightarrow{h^{U_k}_{T_k}}$ and  $\overleftarrow{h^{U_k}_{1}}$ are the last forward and backward utterance-level LSTM states at $k$th turn.

\subsection{Belief Tracking}
    Belief tracking, or dialog state tracking, maintains and adjusts the state of a conversation, such as user's goals, by accumulate evidence along the sequence of a dialog. After collecting new evidence from a user's input at turn $k$, the neural dialog model updates the probability distribution $P(S^{m}_k)$ over candidate values for each slot type $m \in M$. For example, in restaurant search domain, the model maintains a multinomial probability distribution over each of user's goals on restaurant area, food type, and price range. At turn $k$, the dialog-level LSTM ($\operatorname{LSTM_D}$) updates its hidden state $s_k$ and use it to infer any updates on user's goals after taking in the user input encoding $U_k$ and KB indicator $I_k$ (to be described in section below).
        \begin{align}
            & s_k = \operatorname{LSTM_D}(s_{k-1}, \hspace{1mm} [U_k, \hspace{1mm} I_k]) \\
            & P(S^{m}_k \hspace{1mm} | \hspace{1mm} \mathbf{U}_{\le k}, \hspace{1mm} \mathbf{I}_{\le k}) = \operatorname{SlotDist}_{m}(s_k)
        \end{align}
    where $\operatorname{SlotDist}_{m}$ is a multilayer perceptron (MLP) with $\operatorname{softmax}$ activation function over the slot type $m \in M$.

\subsection{Issuing API Calls}
    Conditioning on the state of the conversation, the model may issue an API call to query the KB based on belief tracking outputs. A simple API call command template is firstly generated by the model. The final API call command is produced by replacing the slot type tokens in the command template with the best hypothesis for each of the goal slot from the belief tracker. 
    
    In restaurant search domain, a simple API call command template can be ``$api\_call \hspace{1mm} \left \langle area \right \rangle \hspace{1mm} \left \langle food \right \rangle \hspace{1mm} \left \langle pricerange \right \rangle$'', and the slot type tokens are to be replaced with the belief tracker outputs to form the final API call command ``$api\_call \hspace{1mm} west \hspace{1mm} italian \hspace{1mm} dontcare$''.

\subsection{KB Results Processing}
    Once the neural dialog model receives the KB query results, it suggests options to users by selecting entities from the returned list. Instead of treating KB results as unstructured text (more specifically, as user utterances as in \cite{bordes2016learning,eric2017copy,Seo2017qrn}) and processing them with machine reading comprehension approach, we treat KB results as a list of structured entities and let the model to select appropriate entity pointers. Outputs from KB search or database query typically have well defined structures, with entity attributes associated with entity index. Other than letting the model to learn such entity-attribute association purely from the training dialog corpus as in \cite{bordes2016learning,eric2017copy,Seo2017qrn}, we keep such structural information in our system and let the model to learn to select proper entity pointers from a ranked list. 
 
\begin{comment}   
    We further argue that entity ranking in real world systems can be made with much richer feature sets (e.g. user profiles, location and time context, etc.) in the back end other than just with entity review scores. Therefore, in this work we assume that the neural dialog model receives ranked retrieval results, and the model needs to learn to update entity point based on user's request. 
\end{comment}       

    At turn $k$ of a dialog, a binary KB indicator $\mathbf{I}_{k}$ is passed to the neural dialog model. This indicator is decided by the number of retrieved entities from the last API call and the current entity pointer. When the system is in a state to suggest an entity to user, if a zero value $\mathbf{I}_{k}$ is received, the model is likely to inform user the unavailability of entity matching the current query. Otherwise if $\mathbf{I}_{k}$ has a value of one, the model will likely pick an entity from the retrieved results based on the updated probability distribution of the entity pointer $P(E_k)$:
    \begin{align}
        & P(E_k \hspace{1mm} | \hspace{1mm} \mathbf{U}_{\le k}, \hspace{1mm} \mathbf{I}_{\le k}) = \operatorname{EntityPointerDist}(s_k)
    \end{align}
    where $\operatorname{EntityPointerDist}$ is an MLP with $\operatorname{softmax}$ activation.

\subsection{System Response Generation}
    At $k$th turn of a dialog, a skeletal sentence structure $R_k$ is selected from a list of delexicalised response candidates. The final system response is produced by replacing the delexicalised tokens with the predicted slot values and entity attribute values. For example, replacing $\left \langle food \right \rangle $ to \textit{italian}, and replacing $\left \langle R\_name \right \rangle$ to \textit{prezzo} as in Figure 1.
    \begin{align}
        & P(R_k \hspace{1mm} | \hspace{1mm} \mathbf{U}_{\le k}, \hspace{1mm} \mathbf{I}_{\le k}) = \operatorname{ResponseDist}(s_k)
    \end{align}
    where $\operatorname{ResponseDist}$ is an MLP with $\operatorname{softmax}$ activation.
    
\subsection{Model Training}
    We train the neural dialog model by finding the parameter set $\theta$ that minimize the cross-entropy of the predicted and true distributions for goal slot labels, entity pointer, and delexicalised system response jointly:
        \begin{equation}
            \begin{split}
            \min_{\theta} \sum_{k=1}^{K} -\Big[\hspace{1mm} \sum_{m=1}^{M} &\lambda _{S^{m}} \log P({S^{m}_k}^{*} \hspace{1mm} | \hspace{1mm} \mathbf{U}_{\le k}, \hspace{1mm} \mathbf{I}_{\le k}; \theta) \\
            \hspace{1mm} + \hspace{1mm} &\lambda _E \log P(E_k^{*} \hspace{1mm} | \hspace{1mm} \mathbf{U}_{\le k}, \hspace{1mm} \mathbf{I}_{\le k}; \theta) \\
            \hspace{1mm} + \hspace{1mm} &\lambda _R \log P(R_k^{*} \hspace{1mm} | \hspace{1mm} \mathbf{U}_{\le k}, \hspace{1mm} \mathbf{I}_{\le k}; \theta) \hspace{1mm} \Big] \\
            \end{split}
        \end{equation}
    where $\lambda$s are the linear interpolation weights for the cost of each system output. ${S^{m}_k}^{*}$, $E_k^{*}$, and $R_k^{*}$ are the ground truth labels for each task at the $k$th turn.

\subsection{Alternative Model Designs}     
The model architecture (Figure 1) described above assumes that the hidden state of the dialog-level LSTM  implicitly captures the complete state of the conversation, i.e. the user goal estimation and the previous system actions. Intuitively, the model is likely to provide a better response if it is informed about the goal slot value estimations explicitly and is aware of its previous responses made to the user. Thus, we design and evaluate a few alternative model architectures to verify such assumption:

\textbf{(1)} Model with previously emitted delexicalised system response connected back to dialog-level LSTM state:
        \begin{align}
            s_k = \operatorname{LSTM_D}(s_{k-1}, \hspace{1mm} [U_k, \hspace{1mm} I_k, \hspace{1mm} R_{k-1}])
        \end{align}
        
\textbf{(2)} Model with previously emitted slot labels connected back to dialog-level LSTM state:
        \begin{align}
            s_k = \operatorname{LSTM_D}(s_{k-1}, \hspace{1mm} [U_k, \hspace{1mm} I_k, \hspace{1mm} S^1_{k-1}, ... , S^M_{k-1} ])
        \end{align}
        
\textbf{(3)} Model with both previously emitted response and slot labels connected back to dialog-level LSTM state:
        \begin{align}
             s_k = \operatorname{LSTM_D}(s_{k-1}, \hspace{1mm} [U_k, \hspace{1mm} I_k, \hspace{1mm} R_{k-1}, \hspace{1mm} S^1_{k-1}, ... , S^M_{k-1} ])
        \end{align}

\section{Experiments}
\subsection{Dataset}
    We use data from DSTC2 \cite{henderson2014second} for our model evaluation. This challenge is designed in the restaurant search domain. Bordes and Weston \cite{bordes2016learning} transformed the original DSTC2 corpus by adding system commands and removing the dialog state annotations. This transformed corpus contains additional API calls that the system would make to the KB and the corresponding KB query results. In this study, we combine the original DSTC2 corpus and this transformed version by keeping the dialog state annotations and adding the system commands (API calls). We can thus perform more complete evaluation of our model's capability in tracking the dialog state, processing KB query results, and conducting complete dialog. Statistics of this dataset is summarized in the Table 1.

        \begin{table}[th]
          \caption{Statistics of the converted DSTC2 dataset.}
          \label{tab:table_0}
          \centering
          \begin{tabular}{l r}
            \toprule
            Num of train \& dev / test dialogs           & 2118 / 1117           \\
            Num of turns per dialog in average    & 7.9 \\ 
            \hspace{2mm}(including API call commands)  & \\
            Num of area / food / pricerange options       & 5 / 91/ 3 \\
            Num of delexicalised response candidates & 78 \\
            \bottomrule
          \end{tabular}     
        \end{table}

\subsection{Model Configuration and Training}
    We perform mini-batch model training with batch size of 32 using Adam optimization method \cite{kingma2014adam}. Regularization with dropout is applied to the non-recurrent connections \cite{zaremba2014recurrent} during model training with dropout rate of 0.5. We set the maximum norm for gradient clipping to 5 to prevent exploding gradients.
    
    Hidden layer sizes of the dialog-level LSTM and the utterance-level LSTM are set as 200 and 150 respectively. Word embeddings of size 300 are randomly initialized. We also explore using pre-trained word vectors \cite{mikolov2013distributed} that are trained on Google News dataset to initialize the word embeddings.
    
    \subsection{Results and Analysis}
    Similar to the evaluation methods used in \cite{bordes2016learning,eric2017copy,Seo2017qrn}, we evaluate the task-oriented dialog model in a ranking setting. We report the prediction accuracy for goal slot values, entity pointer, delexicalised system response, and final system response which has the delexicalised tokens replaced by predicted values.

    We first experiment with different text encoding methods and recurrent model architectures to find best performing model. Table 2 shows the evaluation results of models using different user utterance encoding methods and different word embedding initialization. Bidirectional LSTM (Bi-LSTM) shows clear advantage in encoding user utterance comparing to bag-of-means on word embedding (BoW Emb) method, improving the joint goal prediction accuracy by 4.6\% and the final system response accuracy by 1.4\%. Using pre-trained word vectors (word2vec) boosts the model performance further. These results show that the semantic similarities of words captured in the pre-trained word vectors are helpful in generating a better representation of user input, especially when the utterance contains words or entities that are rarely observed during training.

    \begin{table}[th]
      \caption{Prediction accuracy for entity pointer, joint user goal, delexicalised system response, and final system response of the transformed DSTC2 test set using different encoding methods and word vector initialization.}
      \label{tab:table_1}
      \centering
      \begin{tabular}{l c c c c}
        \toprule
         & \textbf{Entity}  & \textbf{Joint} & \textbf{De-lex}  & \textbf{Final} \\
        \textbf{Model} & \textbf{Pointer} & \textbf{Goal} & \textbf{Res}  & \textbf{Res} \\
        \midrule
        BoW Emb Encoder                      & 93.5 & 72.6 & 55.4 & 51.2 \\
        \hspace{5mm} + word2vec          & 93.6 & 74.3 & 55.9 & 51.5 \\
        Bi-LSTM Encoder                    & 93.8 & \textbf{77.2} & 55.8 & 52.6 \\
        \hspace{5mm} + word2vec          & \textbf{94.4} & 76.6 & \textbf{56.6} & \textbf{52.8} \\
        \bottomrule
      \end{tabular}
    \end{table}

    Table 3 shows the evaluation results of different recurrent model architectures. The Hierarchical LSTM model in Table 3 refers to the last model in Table 2. Models in row 2 to 4 of Table 3 refer to the three recurrent model architectures discussed in section 3.7. As illustrated in these results, models that conditioned on previous emitted labels in generating system response achieve lower prediction accuracy across all of the four evaluation metrics. These observations are contrary to our intuition and analysis made in previous section. We believe the degraded performance is mainly due to the data sparsity issue of the dataset that used in the experiment. Given a certain dialog context, there might be multiple system response candidates that can be used to generate a suitable final response. With limited number of training samples, the model is likely to overfit the training set and not to generalize the strong modeling capacity well during inference. The overfitting issue might be less of a problem in the word-by-word response generation setting, and this is to be studied further in our future work.
    
    \begin{table}[th]
      \caption{Prediction accuracy with different recurrent model architectures.}
      \label{tab:table_2}
      \centering
      \begin{tabular}{l c c c c}
        \toprule
         & \textbf{Entity}  & \textbf{Joint} & \textbf{De-lex}  & \textbf{Final} \\
        \textbf{Model} & \textbf{Pointer} & \textbf{Goal} & \textbf{Res}  & \textbf{Res} \\
        \midrule
        Hierarchical LSTM      & \textbf{94.4} & \textbf{76.6} & \textbf{56.6} & \textbf{52.8} \\
        \hspace{1mm} + feed de-lex res (1)      & 93.6 & 74.8 & 55.4 & 51.8 \\
        \hspace{1mm} + feed goal slots (2)      & 94.1 & 75.3 & 55.3 & 51.8 \\
        \hspace{1mm} + feed both  (3)   & 93.7 & 72.7 & 55.3 & 51.6 \\
        \bottomrule
      \end{tabular}
    \end{table}
    
   As observed in Table 3, the joint goal tracking performance is directly related to final response prediction accuracy. To further investigate our model's capability on belief tracking, we conduct error breakdown analysis for each goal slot. We compare our model to two other recently proposed belief tracking models, an RNN based model \cite{henderson2014robust} and the Neural Belief Tracker \cite{mrkvsic2016neural}, in the setting of only using live ASR hypothesis as model input. As the results show in Table 4, our system achieves promising belief tracking performance comparable to the state-of-the-art systems. 
   
   % As shown in the results in Table 4, tracking user's goal for \textit{food} is the most challenging one among the three goal tracking tasks. Comparing to other belief tracking models, our proposed method shows advantageous performance in tracking user's goal for \textit{food}, and similar performance in tracking user's goal for \textit{area} and \textit{price}. Our model's superior performance in \textit{food} goal tracking directly contributes to the improved joint goal prediction accuracy.

    \begin{table}[th]
      \caption{Dialog state tracking performance on DSTC2 test set, comparing to previous approaches.}
      \label{tab:table_3}
      \centering
      \begin{tabular}{l c c c c}
        \toprule  
        & \textbf{Area}  & \textbf{Food}  & \textbf{Price}  & \textbf{Joint} \\
         \textbf{Model} &   \textbf{Goal}  & \textbf{Goal}  & \textbf{Goal}  & \textbf{Goal} \\
        \midrule
        RNN    & 92 & 86 & 86 & 69  \\
        RNN + sem. dict    & 91 & 86 & 93 & 73  \\
        NBT-DNN \cite{mrkvsic2016neural}      & 90 & 84 & 94 & 72  \\
        NBT-CNN \cite{mrkvsic2016neural}      & 90 & 83 & 93 & 72  \\
        \midrule
        Hierarchical LSTM      & 90 & 84 & 93 & 73 \\
        \bottomrule
      \end{tabular}     
    \end{table}

    % Finally, we compare the performance of the proposed method to related published works, using the same per-response accuracy metric. 
    Finally, we report the model performance in producing final system responses and compare it to other published results following the per-response accuracy metric used in prior work. Even though using the same evaluation measurement, our model is designed with slightly different settings comparing to other published models in Table 5. Instead of using additional matched type features (i.e. KB entity type feature for each word, e.g. whether a word is a food type or area type, etc.) as in \cite{bordes2016learning,Seo2017qrn}, we use user's goal slots at each turn that are mapped from the original DSTC2 dataset as additional supervised signals in our model. Moreover, instead of treating KB query results as unstructured text, we treat them as structured entities and let our model to pick the right entity by selecting the most appropriate entity pointer. 
    %Although these per-response prediction results may not be directly comparable, they give hints on the models' capacity in conducting task-oriented dialogs.} 
    Our proposed model successfully predicts 52.8\% of the true system responses, outperforming prior end-to-end trainable neural dialog systems.

    \begin{table}[th]
      \caption{Performance of the proposed model in per-response accuracy comparing to previous approaches.}
      \label{tab:table_4}
      \centering
      \begin{tabular}{l c}
        \toprule
        \textbf{Model} & \textbf{Per-res Accuracy}\\
        \midrule
        Memory Networks \cite{bordes2016learning}      & 41.1 \\
        Gated Memory Networks \cite{perez2016gated}     & 48.7 \\
        Sequence-to-Sequence \cite{eric2017copy}    & 48.0 \\
        Query-Reduction Networks \cite{Seo2017qrn}        & 51.1 \\
        \midrule
        Hierarchical LSTM & \textbf{52.8} \\
        \bottomrule
      \end{tabular}
    \end{table}
    
    To further understand the prediction errors made by our model, we conduct human evaluation by inviting 10 users to evaluate the appropriateness of the responses generated by our system. While some of the errors are made on generating proper API calls due to the errors in dialog state tracking results, we also find quite a number of responses that are considered appropriate by our judges but do not match to the reference responses in the test set. For example, there are cases where our system directly issues the correct API call (e.g. ``$api\_call \hspace{1mm} south \hspace{1mm} italian \hspace{1mm} expensive$'') based on user's inputs, instead of asking user for confirmation of a goal type  (e.g. "\textit{Did you say you are looking for a restaurant in the south of town?}") as in the reference corpus. By taking such factors into consideration, our system was able to generate appropriate responses in 73.6\% of the time based on feedback from the judges. These results show that the per-response accuracy evaluation metric may not well correlate with human judgments \cite{liu2016not}, and better dialogs evaluation measurements should to be further explored.

\section{Conclusions}
In this work, we propose a novel end-to-end trainable neural network model for task-oriented dialog systems. The model is able to track dialog belief state, interface with knowledge bases by issuing API calls, and incorporate structured query results into system responses to successfully complete task-oriented dialogs. In the evaluation in a restaurant search domain using a converted dataset from the second Dialog State Tracking Challenge corpus, our proposed model shows robust performance in tracking dialog state over the sequence of dialog turns. The model also demonstrates promising performance in generating appropriate system responses, outperforming prior end-to-end trainable neural network models.

\bibliographystyle{IEEEtran}

\bibliography{mybib}

\end{document}